\documentclass[10pt, a4paper]{article}
\usepackage{lrec2014}
\usepackage{graphicx}

\usepackage{epstopdf}
\usepackage[latin1]{inputenc}

\usepackage{arydshln}
\usepackage{amsmath}
\usepackage{multirow}
\usepackage{graphicx}
\usepackage{color}
\usepackage{times}
\usepackage{latexsym} 

\title{\textsc{DefExt}: A Semi Supervised Definition Extraction Tool}

\name{Luis Espinosa-Anke, Roberto Carlini, Horacio Saggion, Francesco Ronzano}

\address{ TALN Group, Universitat Pompeu Fabra \\
               Carrer T\`{a}nger, 122-134, Barcelona (Spain) \\
               \{luis.espinosa,firstname.lastname\}@upf.edu\\}

\abstract{We present \textsc{DefExt}, an easy to use semi supervised Definition Extraction Tool. \textsc{DefExt} is designed to extract from a target corpus those textual fragments where a term is explicitly mentioned together with its core features, i.e. its definition. It works on the back of a Conditional Random Fields based sequential labeling algorithm and a bootstrapping approach. Bootstrapping enables the model to gradually become more aware of the idiosyncrasies of the target corpus. In this paper we describe the main components of the toolkit as well as experimental results stemming from both automatic and manual evaluation. We release \textsc{DefExt} as open source along with the necessary files to run it in any Unix machine. We also provide access to training and test data for immediate use.
\\ \newline }

\begin{document}

\maketitleabstract

\section{Introduction}

Definitions are the source of knowledge to consult when the meaning of a term is sought, but manually constructing and updating glossaries is a costly task which requires the cooperative effort of domain experts \cite{NavigliandVelardi2010}. Exploiting lexicographic information in the form of definitions has proven useful not only for Glossary Building \cite{MuresanandKlavans2002,Parketal2002} or Question Answering \cite{Cuietal2005,SaggionandGaizauskas2004}, but also more recently in tasks like Hypernym Extraction \cite{Espinosa-Ankeetal2015}, Taxonomy Learning \cite{Velardietal2013,Espinosa-Ankeetal2016} and Knowledge Base Generation \cite{DelliBovietal2015b}.

Definition Extraction (DE), i.e. the task to automatically extract definitions from naturally occurring text, can be approached by exploiting lexico-syntactic patterns \cite{RebeyrolleandTanguy2000,Sarmentoetoetal2006,StorrerandWellinghoff2006}, in a supervised machine learning setting \cite{Naviglietal2010,Yipingetal2013,Espinosa-AnkeandSaggion2014,Espinosa-Ankeetal2016SEPLN}, or leveraging bootstrapping algorithms \cite{Reiplingeretal2012,DeBenedictisetal2013}.

In this paper, we extend our most recent contribution to DE by releasing \textsc{DefExt}\footnote{https://bitbucket.org/luisespinosa/defext}, a toolkit based on experiments described in \cite{Espinosa-Ankeetal2015b}, consisting in machine learning sentence-level DE along with a bootstrapping approach. First, we provide a summary of the foundational components of \textsc{DefExt} (Section \ref{sec:datamodelling}). Next, we summarize the contribution from which it stems \cite{Espinosa-Ankeetal2015b}  as well as its main conclusion, namely that our approach effectively generates a model that gradually adapts to a target domain (Section \ref{sec:defexteval}). Furthermore, we introduce one additional evaluation where, after bootstrapping a subset of the ACL Anthology, we present human experts in NLP with definitions and distractors (Section \ref{sec:manualeval}), and ask them to judge whether the sentence includes \textit{definitional knowledge}. Finally, we provide a brief description of the released toolkit along with accompanying enriched corpora to enable immediate use (Section \ref{sec:techdetails}).

\section{Data Modelling}
\label{sec:datamodelling}

\begin{figure}[h!]
  \centering
    \includegraphics[height=4cm,width=8cm]{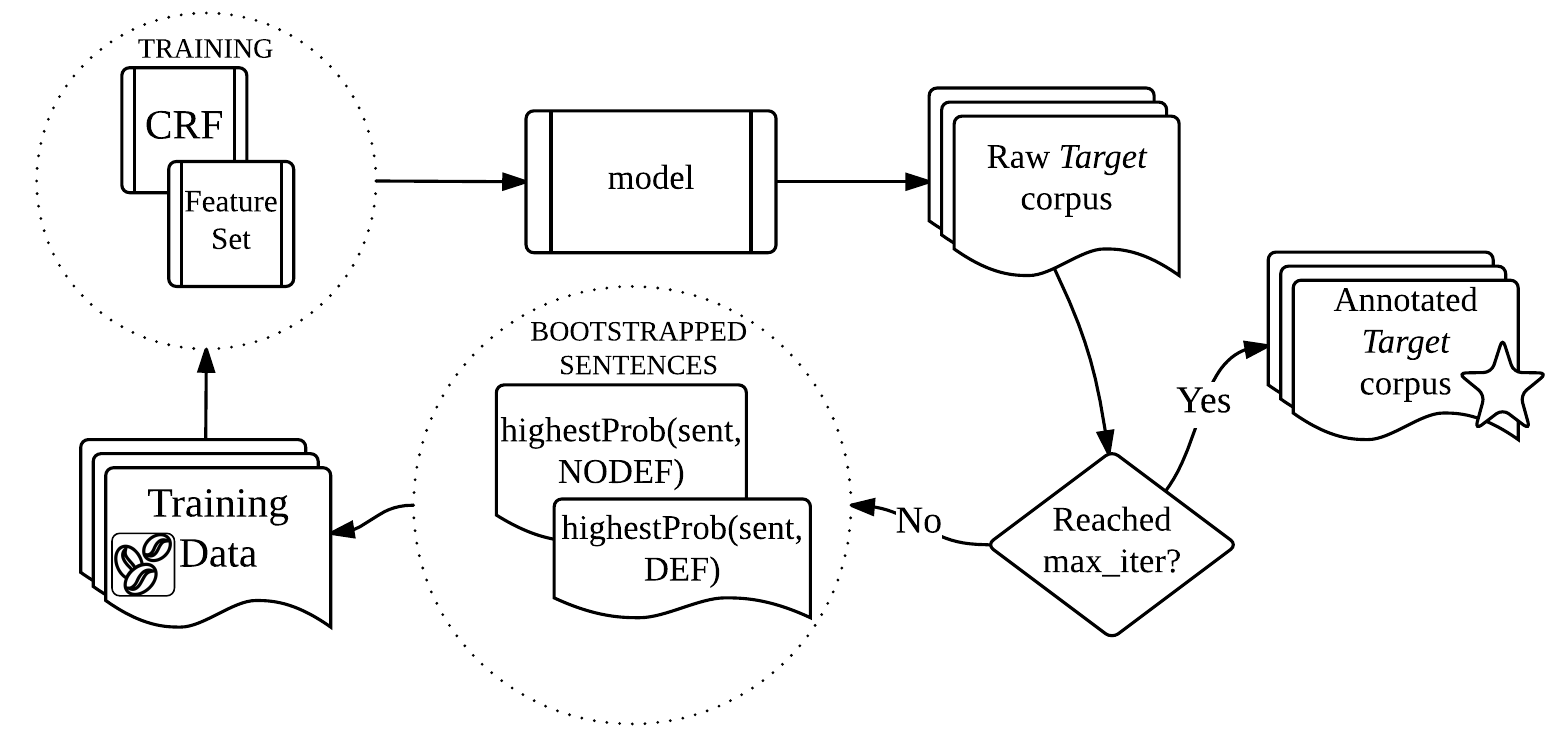}
  \caption{Workflow of \textsc{DefExt}.}
  \label{fig:architecture}
\end{figure}

\textsc{DefExt} is a weakly supervised DE system based on Conditional Random Fields (CRF) which, starting from a set of manually validated definitions and distractors, trains a seed model and iteratively enriches it with high confidence instances (i.e. highly likely definition sentences, and highly likely not definition sentences, such as a text fragments expressing a personal opinion). What differentiates \textsc{DefExt} from any supervised system is that thanks to its iterative architecture (see Figure \ref{fig:architecture}), it gradually identifies more appropriate definitions with larger coverage on non-standard text than a system trained only on WordNet glosses or Wikipedia definitions (experimental results supporting this claim are briefly discussed in Section \ref{sec:ranlpexperiment}).

\subsection{Corpora}

We release \textsc{DefExt} along with two automatically annotated datasets in column format, where each line represents a word and each column a feature value, the last column being reserved for the token's label, which in our setting is \textsc{DEF}/\textsc{NODEF}. Sentence splitting is encoded as a double line break.

These corpora are enriched versions of (1) The Word Class Lattices corpus \cite{Naviglietal2010} (WCL$_d$); and (2) A subset of the ACL Anthology Reference Corpus \cite{Birdetal2008} (ACL-ARC$_d$). Statistics about their size in tokens and sentences, as well as definitional distribution (in the case of WCL$_d$) are provided in Table \ref{tab:datastats}.

\begin{table}[]
\centering
\begin{tabular}{lrr}
\multicolumn{1}{c}{} & \multicolumn{1}{c}{\textbf{WCL$_d$}} & \multicolumn{1}{c}{\textbf{ACL-ARC$_d$}} \\ \hline
Sentences   & 3,707   & 241,383          \\ \hline
Tokens      & 103,037 & 6,709,314 \\ \hline
Definitions & 2,059   & NA         \\ \hline
Distractors & 1,644   & NA       
\end{tabular}
\caption{Statistics of the two enriched corpora that accompany the \textsc{DefExt} code.}
\label{tab:datastats}
\end{table}

\subsection{Feature Extraction}

The task of DE is modelled as a sentence classification problem, i.e. a sentence may or may not contain a definition. The opting for CRF as the machine learning algorithm is twofold: First, its sequential nature allows us to encode fine-grained features at word level, also considering the context of each word. And second, it has proven useful in previous work in the task of DE \cite{Yipingetal2013}.

The features used for modelling the data are based on both linguistic, lexicographic and statistical information, such as:

\begin{itemize}
    \item \textbf{Linguistic Features}: Surface and lemmatized words, part-of-speech, chunking information (NPs) and syntactic dependencies.
    \item \textbf{Lexicographic Information}: A feature that looks at noun phrases and whether they appear at potential \textit{definiendum} ($D$) or \textit{definiens} ($d$) position\footnote{The \textit{genus et differentia} model of a definition, which traces back to Aristotelian times, distinguishes between the \textit{definiendum}, the term that is being defined, and the \textit{definiens}, i.e. the cluster of words that describe the core characteristics of the term.}, as illustrated in the following example:\\  The$\langle$o-D$\rangle$ Abwehr$\langle$b-D$\rangle$ was$\langle$o-d$\rangle$ a$\langle$o-d$\rangle$ German$\langle$b-d$\rangle$ intelligence$\langle$i-d$\rangle$ organization$\langle$i-d$\rangle$ from$\langle$o-d$\rangle$ 1921$\langle$o-d$\rangle$ to$\langle$o-d$\rangle$ 1944$\langle$o-d$\rangle$.
    \item \textbf{Statistical Features}: These are features designed to capture the degree of \textit{termhood} of a word, its frequency in generic or domain-specific corpora, or evidence of their salience in definitional knowledge. These are:
    
    \begin{itemize}
        \item \textbf{termhood}: This metric determines the importance of a candidate token to be a terminological unit by looking at its frequency in general and domain-specific corpora \cite{KitandLiu2008}. It is obtained as follows:

$$\text{\mbox{Termhood}(w)} = \frac{ \text{r}_\text{D}\text{(w)} }{|\text{\mbox{V}}_\text{D}|}-\frac{ \text{r}_\text{B}\text{(w)} }{|\text{\mbox{V}}_\text{B}|} $$

Where $\text{r}_\text{D}$ is the frequency-wise ranking of word $w$ in a domain corpus (in our case, WCL$_d$), and $\text{r}_\text{B}$ is the frequency-wise ranking of such word in a general corpus, namely the Brown corpus \cite{Brown1979}. Denominators refer to the token-level size of each corpus. If word $w$ only appears in the general corpus, we set the value of Termhood(w) to $- \infty$, and to $\infty$ in the opposite case.

\item \textbf{tf-gen}: Frequency of the current word in the general-domain corpus $\text{r}_\text{B}$ (Brown Corpus).
\item \textbf{tf-dom}: Frequency of the current word in the domain-specific corpus $\text{r}_\text{D}$ (WCL$_d$).
\item \textbf{tfidf}: Tf-idf of the current word over the training set, where each sentence is considered a separate document.
\item \textbf{def\_prom}: The notion of Definitional Prominence describes the probability of a word $w$ to appear in a definitional sentence ($s = def$). For this, we consider its frequency in definitions and non-definitions in the WCL$_d$ as follows:

$$\text{\mbox{DefProm}(w)} = 
\frac{\text{DF}}
	{|\text{Defs}|} - 
\frac{ \text{NF} }
	{|\text{Nodefs}|} $$

where $\text{DF} = \sum_{i=0}^{i=n} (s_i = $\emph{ def} $\wedge$ $w \in s_i)$ and $\text{NF} = \sum_{i=0}^{i=n} (s_i = $\emph{ nodef} $\wedge$ $w \in s_i)$. Similarly as with the \textit{termhood} feature, in cases where a word $w$ is only found in definitional sentences, we set the DefProm(w) value to $\infty$, and to $- \infty$ if it was only seen in non-definitional sentences.

\item \textbf{D\_prom}: Definiendum Prominence, on the other hand, models our intuition that a word appearing more often in position of potential \textit{definiendum} might reveal its role as a definitional keyword. This feature is computed as follows: 

$$
\text{\text{DP}(w)} = \frac{\sum_{i=0}^{i=n}w_i\in\mbox{term}_D}{|DT|}
$$

where $\mbox{term}_D$ is a noun phrase (i.e. a term candidate) appearing in potential definiendum position and $|\mbox{DT}|$ refers to the size of the candidate term corpus in candidate definienda position.

\item \textbf{d\_prom}: Similarly computed as D\_prom, but considering position of potential definiens.

    \end{itemize}
\end{itemize}

These features are used to train a CRF algorithm. \textsc{DefExt} operates on the back of the CRF toolkit \textsc{CRF++}\footnote{https://taku910.github.io/crfpp/}, which allows selecting features to be considered at each iteration, as well as the context window.

\subsection{Bootstrapping}

We implemented on \textsc{DefExt} a bootstrapping approach inspired by the well-known Yarowsky algorithm for Word Sense Disambiguation \cite{Yarowsky1995}. It works as follows: Assuming a small set of seed labeled examples(in our case, WCL$_d$), a large target dataset of cases to be classified (ACL-ARC$_d$), and a learning algorithm (CRF), the initial training is performed on the initial seeds in order to classify the whole target data. Those instances classified with high confidence are appended to the training data until convergence or a number of maximum iterations is reached.

We apply this methodology to the definition bootstrapping process, and for each iteration, extract the highest confidence definition and the highest confidence non-definition from the target corpus, retrain, and classify again. The number of maximum iterations may be introduced as an input parameter by the user. Only the latest versions of each corpus are kept in disk.

\section{Experiments and Evaluation}
\label{sec:ranlpexperiment}

As the bootstrapping process advances, the trained models gradually become more aware of the linguistic particularities of the genre and register of a target corpus. This allows capturing definition fragments with a particular syntactic structure which may not exist in the original seeds. In this section, we summarize the main conclusions drawn from the experiments performed over two held out test datasets, namely the W00 corpus \cite{Yipingetal2013}, and the MSR-NLP corpus \cite{Espinosa-Ankeetal2015b}\footnote{Available at http://taln.upf.edu/MSR-NLP\_RANLP2015}. The former is a manually annotated subset of the ACL anthology, which shows high domain-specifity as well as considerable variability in terms of how a term is introduced and defined (e.g. by means of a comparison or by placing the defined term at the end of the sentence). The latter is compiled manually from a set of abstracts available at the Microsoft Research website\footnote{http://academic.research.microsoft.com}, where the first sentence of each abstract is tagged in the website as a definition. This corpus shows less linguistic variability and thus its definitions are in the vast majority of cases, highly canonical. We show one sample definition from each corpus in Table \ref{tab:examples}\footnote{Note that, for clarity, we have removed from the examples any metainformation present in the original datasets.}.

\begin{table*}[]
\centering
\def\arraystretch{1.5}
\setlength{\tabcolsep}{.1cm}
\begin{tabular}{l|l}
\multicolumn{1}{c|}{\textbf{\textsc{Corpus}}} & \multicolumn{1}{c}{\textbf{\textsc{Definition}}}                                                                         \\ \hline \hline
\textbf{WCL$_d$}                                  & \begin{tabular}[c]{@{}l@{}}The Abwehr was a German intelligence organization from 1921 to 1944 .\end{tabular} 
\\
\hline
\textbf{W00}                                  & \begin{tabular}[c]{@{}l@{}}Discourse refers to any form of language-based communication involving multiple\\ sentences or utterances.  \end{tabular}
\\

\hline
\textbf{ACL-ARC$_d$}                                  & \begin{tabular}[c]{@{}l@{}} In computational linguistics, word sense disambiguation (WSD) is an open problem of natural\\ language processing, which governs the process of identifying which sense of a word.\end{tabular} \\

\hline
\textbf{MSR-NLP}                              & \begin{tabular}[c]{@{}l@{}}User interface is a way by which a user communicates with a computer through a particular\\ software application.\end{tabular}
\\
\end{tabular}
\caption{Example definitions of all corpora involved in the development and evaluation of \textsc{DefExt}.}
\label{tab:examples}
\end{table*}

\subsection{Definition Extraction}
\label{sec:defexteval}


Starting with the WCL$_d$ corpus as seed data, and the ACL-ARC$_d$ collection for bootstrapping, we performed 200 iterations and, at every iteration, we computed Precision, Recall and F-Score at sentence level for both the W00 and the MSR-NLP corpora. At iteration 100, we recalculated the statistical features over the bootstrapped training data (which included 200 more sentences, 100 of each label). 

\def\arraystretch{1.5}
\begin{table}[!t]
\centering
\begin{tabular}{l|cccc}
\multicolumn{1}{l|}{} & Iteration & P & R & F \\ \hline\hline
MSR-NLP & 20 & 78.2 & 76.7 & 77.44 \\ 
W00 & 198 & 62.47 & 82.01 & 71.85 \\ 
\end{tabular}
\caption{Iteration and best results for the two held-out test datasets on the DE experiment.}
\label{tab:best}
\end{table}

\begin{figure}    
\begin{minipage}[]{0.2\textwidth}
\includegraphics[width=3.75cm,height=3cm]{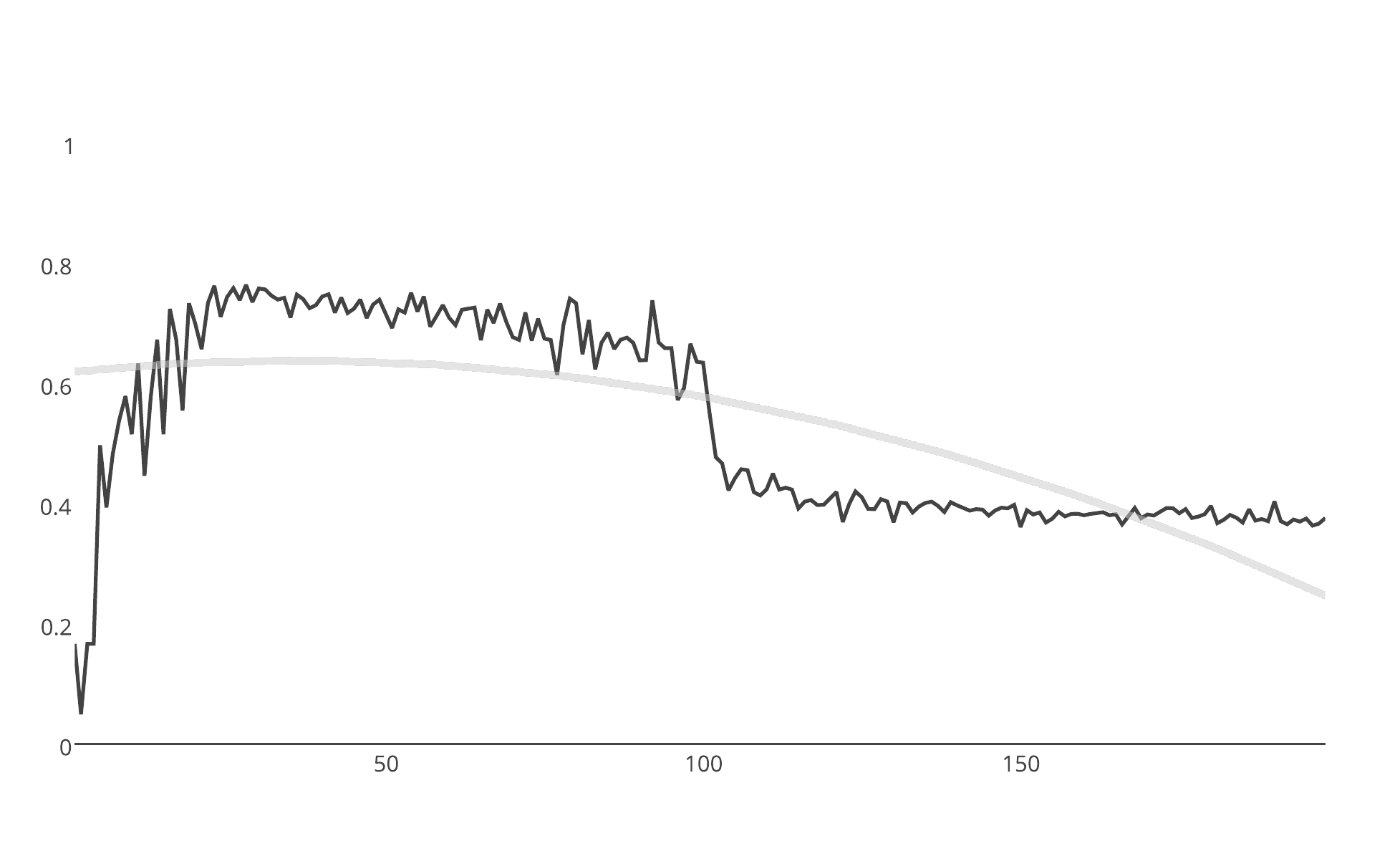}
\end{minipage}
\hspace*{2em}
\begin{minipage}[]{0.2\textwidth}
\includegraphics[width=3.75cm,height=3cm]{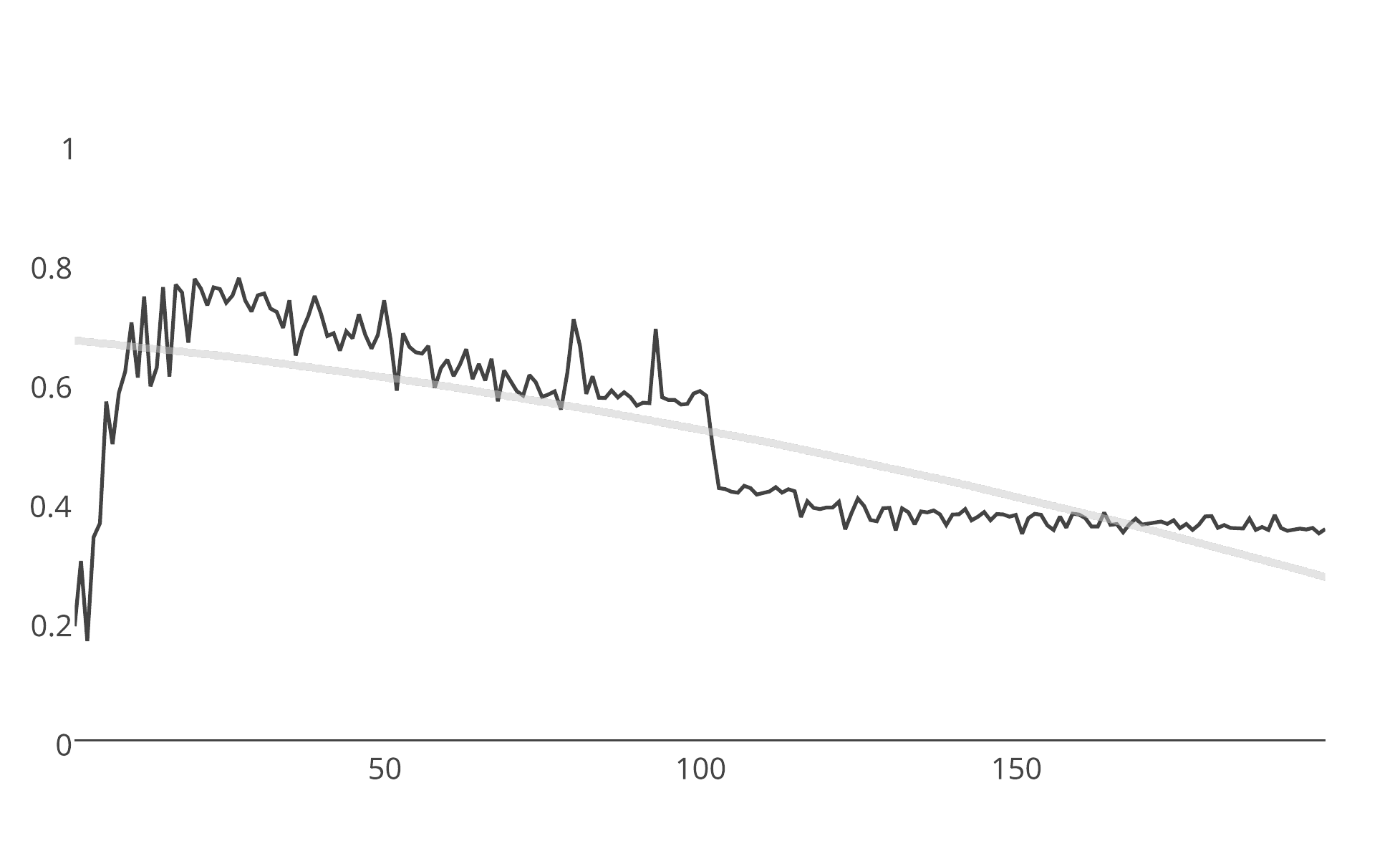}
\end{minipage}

\vspace*{0.2cm} 
\begin{minipage}[]{0.2\textwidth}
\includegraphics[width=3.75cm,height=3cm]{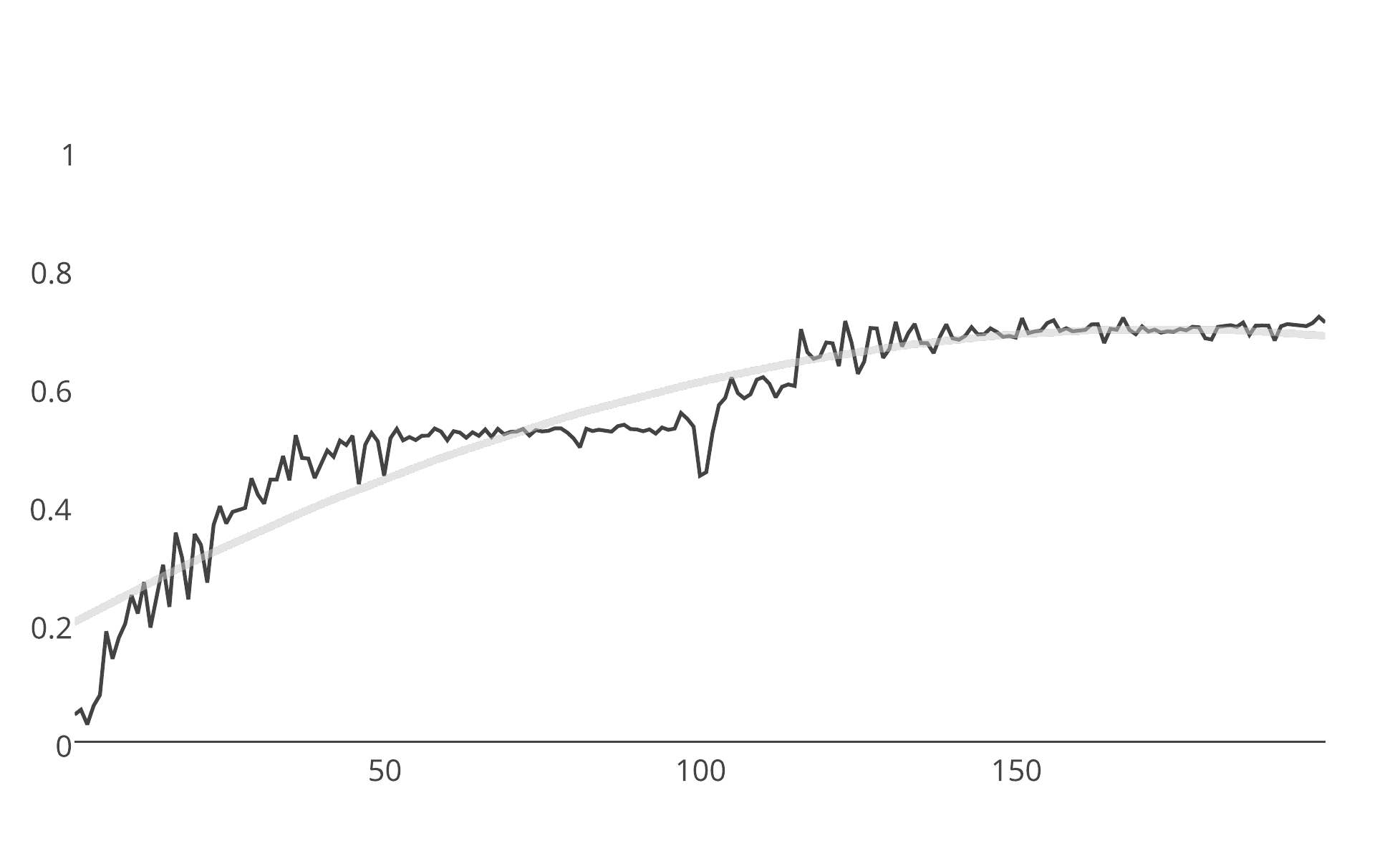}
\end{minipage}
\hspace*{2em}
\begin{minipage}[]{0.2\textwidth}
\includegraphics[width=3.75cm,height=3cm]{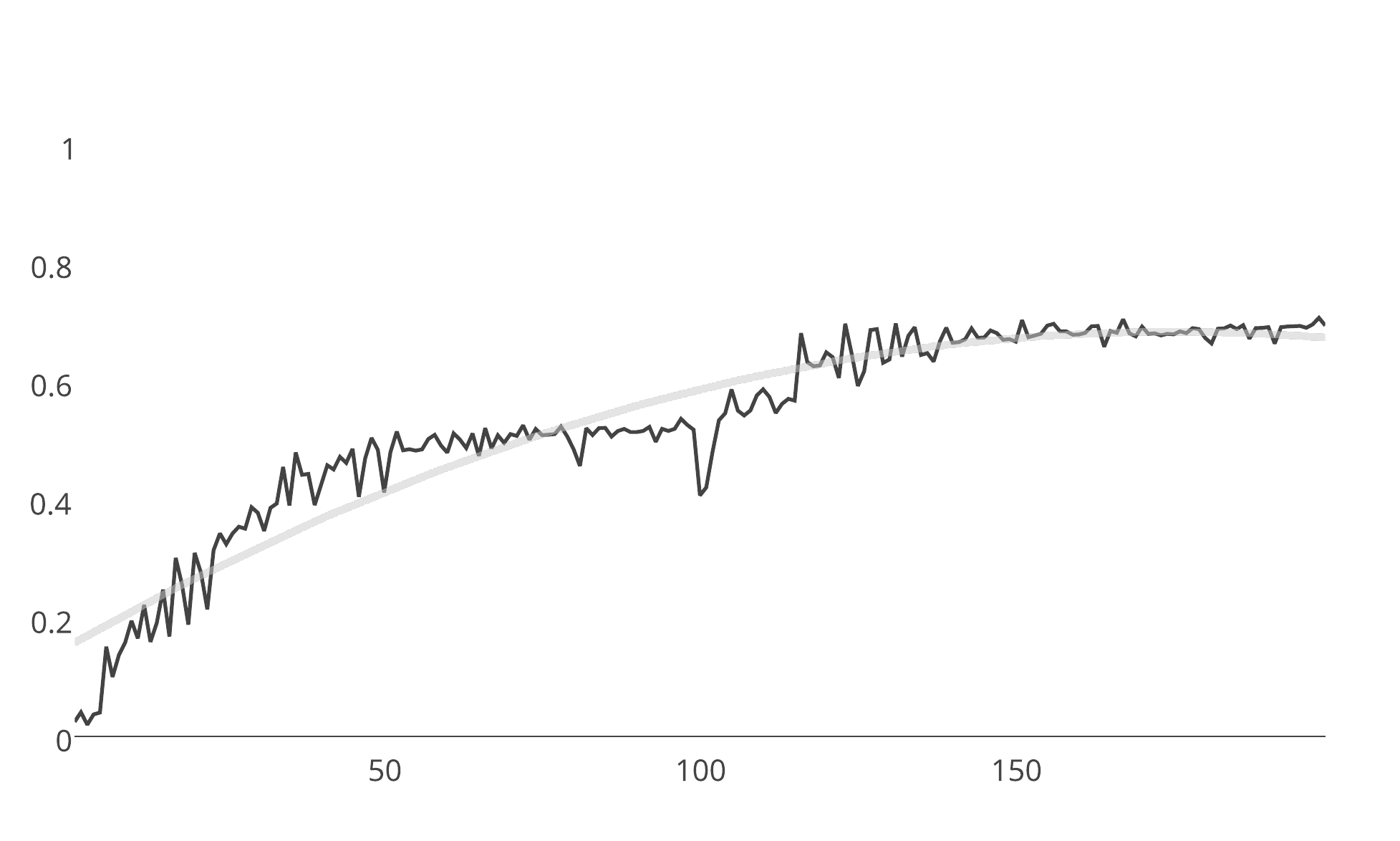}
\end{minipage}
\caption{F-Score against iteration on the MSR-NLP (top row) and W00 (bottom row) datasets under two different confidence thresholds.}
\label{fig:results}
\end{figure}

The trend for both corpora, shown in Figure \ref{fig:results}, indicates that the model improves at classifying definitional knowledge in corpora with greater variability, as performance on the W00 corpus suggests. Moreover, it shows decreasing performance on standard language. The best iterations with their corresponding scores for both datasets are shown in Table \ref{tab:best}.

\subsection{Human Evaluation}
\label{sec:manualeval}

In this additional experiment, we assessed the quality of extracted definitions during the bootstrapping process. To this end, we performed 100 iterations over the ACL-ARC$_d$ corpus and presented experts in NLP with 200 sentences (100 candidate definitions and 100 distractors), with shuffled order. Note that since the ACL-ARC corpus comes from parsing pdf papers, there is a considerable amount of noise derived from diverse formatting, presence of equations or tables, and so on. All sentences, however, were presented in their original form, noise included, as in many cases we found that even noise could give the reader an idea of the context in which the sentence was uttered (e.g. if it is followed by a formula, or if it points to a figure or table).

The experiment was completed by two judges who had extensive familiarity with the NLP domain and its terminology. Evaluators were allowed to leave the answer field blank if the sentence was unreadable due to noise. 

In this experiment, \textsc{DefExt} reached an average (over the scores provided by both judges) of 0.50 Precision when computed over the whole dataset, and 0.65 if we only consider sentences which were not considered noise by the evaluators. Evaluators found an average of 23 sentences that they considered unreadable.


\subsection{Technical Details}
\label{sec:techdetails}

As mentioned earlier, \textsc{DefExt} is a bootstrapping wrapper around the CRF toolkit \textsc{CRF++}, which requires input data to be preprocessed in column-based format. Specifically, each sentence is encoded as a matrix, where each word is a row and each feature is represented as a column, each of them tab-separated. Usually, the word's surface form or lemma will be at the first or second column, and then other features such as part-of-speech, syntactic dependency or corpus-based features follow. 

%
The last column in the dataset is the sentence label, which in \textsc{DefExt} is either \textsc{DEF} or \textsc{NODEF}, as it is designed as a sentence classification system. 


%
Once training and target data are preprocessed accordingly, one may simply invoke \textsc{DefExt} in any Unix machine. Further implementation details and command line arguments can be found in the toolkit's documentation, as well as in comments throughout the code.

\section{Discussion and Conclusion}

We have presented \textsc{DefExt}, a system for weakly supervised DE at sentence level. We have summarized the most outstanding features of the algorithm by referring to experiments which took the NLP domain as a use case \cite{Espinosa-Ankeetal2015b}, and complemented them with one additional human evaluation. We have also covered the main requirements for it to function properly, such as data format and command line arguments. No external Python libraries are required, and the only prerequisite is to have \textsc{CRF++} installed. We hope the research community in lexicography, computational lexicography or corpus linguistics find this tool useful for automating term and definition extraction, for example, as a support for glossary generation or hypernymic (is-a) relation extraction.

\section{Acknowledgements}

This work is partially funded by the Spanish Ministry of Economy and Competitiveness under the Maria de Maeztu Units of Excellence Programme (MDM-2015-0502) and Dr. Inventor (FP7-ICT-2013.8.1611383).

\bibliographystyle{lrec2014}
\bibliography{thesisbib}

\end{document}